\documentclass{article}

\PassOptionsToPackage{numbers,compress}{natbib}
\usepackage[preprint]{neurips_2026}

\usepackage[utf8]{inputenc}
\usepackage[T1]{fontenc}
\usepackage{courier}
\usepackage{hyperref}
\usepackage{url}
\usepackage{booktabs}
\usepackage{microtype}
\usepackage[table]{xcolor}
\usepackage{tabularx}
\usepackage{array}
\usepackage{multirow}
\usepackage{graphicx}
\usepackage{subcaption}
\usepackage{amsmath}
\usepackage{amssymb}
\usepackage{enumitem}
\usepackage{pifont}
\usepackage[capitalise,nameinlink]{cleveref}
\usepackage{tikz}
\newcommand{\benchmark}{\textbf{CUDAHercules}}
\newcommand{\taskyaml}{\texttt{task.yaml}}
\newcommand{\fastp}{\texttt{fast\_p}}
\definecolor{goodgreen}{RGB}{34,139,34}
\definecolor{badred}{RGB}{180,38,38}
\newcommand{\cmark}{\textcolor{goodgreen}{\ding{51}}}
\newcommand{\xmark}{\textcolor{badred}{\ding{55}}}

\newcommand{\numtasks}{195}
\newcommand{\numclassone}{63}
\newcommand{\numclasstwo}{119}
\newcommand{\numclassthree}{10}
\newcommand{\numclassfour}{3}
\newcommand{\cOneGeneral}{20}
\newcommand{\cOneHopper}{21}
\newcommand{\cOneBlackwell}{22}
\newcommand{\cTwoGeneral}{43}
\newcommand{\cTwoHopper}{64}
\newcommand{\cTwoBlackwell}{12}

\title{CUDAHercules\,\smash{\raisebox{-0.35em}{\includegraphics[height=1.75em]{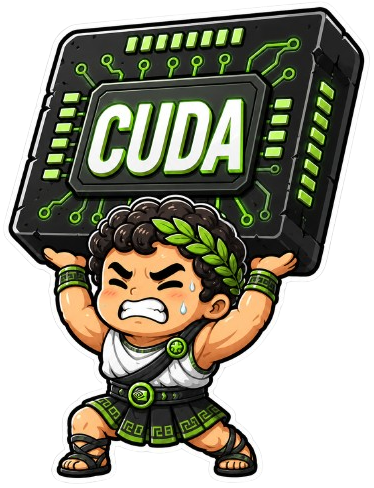}}}: Benchmarking Hardware-Aware Expert-level CUDA Optimization for LLMs}
\author{%
  Shiyang Li \\
  University of Minnesota \\
  \texttt{li004074@umn.edu}
  \And
   Zijian Zhang \\
  University of Minnesota\\
  \texttt{zha00175@umn.edu}
  \And
  Guangyan Sun \\
  University of Minnesota \\
  \texttt{sun01158@umn.edu}
  \And
  Yuebo Luo \\
  University of Minnesota \\
  \texttt{luo00466@umn.edu}
  \And
  Winson Chen \\
  University of Minnesota \\
  \texttt{chen9619@umn.edu}
  \And
  Yanzhi Wang \\
  Northeastern University \\
  \texttt{yanz.wang@northeastern.edu}
  \And
  Mingyi Hong \\
  University of Minnesota \\
  \texttt{mhong@umn.edu}
  \And
  Caiwen Ding \\
  University of Minnesota \\
  \texttt{dingc@umn.edu}
  \And
}
\date{}

\begin{document}
\maketitle

\begin{abstract}
Large language models show promise for automated CUDA programming, however even the strongest coding models (e.g., Claude-Opus-4.6) may still fall short of expert-level, architecture-aware optimization. We introduce \benchmark, a benchmark that evaluates generated CUDA against end-to-end human-expert SOTA systems. It spans single kernels, module-level operators, full applications, and unsolved challenge tasks across Ampere, Hopper, and Blackwell GPUs, with end-to-end tasks gated by domain-specific semantic validators. Evaluating models such as Claude-Opus-4.6 and GPT-5.4 shows a large gap between runnable CUDA and expert CUDA engineering: models often compile and pass tests, but rarely recover the optimization strategies needed to match expert performance. Application semantics further reduce success, and iterative or tool-augmented feedback can improve correctness while drifting toward slow fallback implementations. These results show that automated CUDA programming remains far from fully solved and requires stronger hardware reasoning, better tool use, and training objectives that connect code understanding to hardware architecture-grounded intelligence.
\end{abstract}

\section{Introduction}

Large language model (LLM)-based systems have recently shown substantial potential for automated CUDA programming~\cite{avoagenticvariationoperators,cudaforge,drkernel,stitchcuda,kevin32b}. Given a PyTorch operator~\cite{ouyang2025kernelbench,li2025tritonbench} or natural-language specification~\cite{computeeval2025}, current models can often generate compilable CUDA code and sometimes outperform functional baseline implementations. This progress raises an important question for GPU systems research: can LLMs move beyond functional code generation and perform the kind of deep optimization required in expert CUDA engineering?

\begin{figure}[t]
    \centering
    \includegraphics[width=\linewidth]{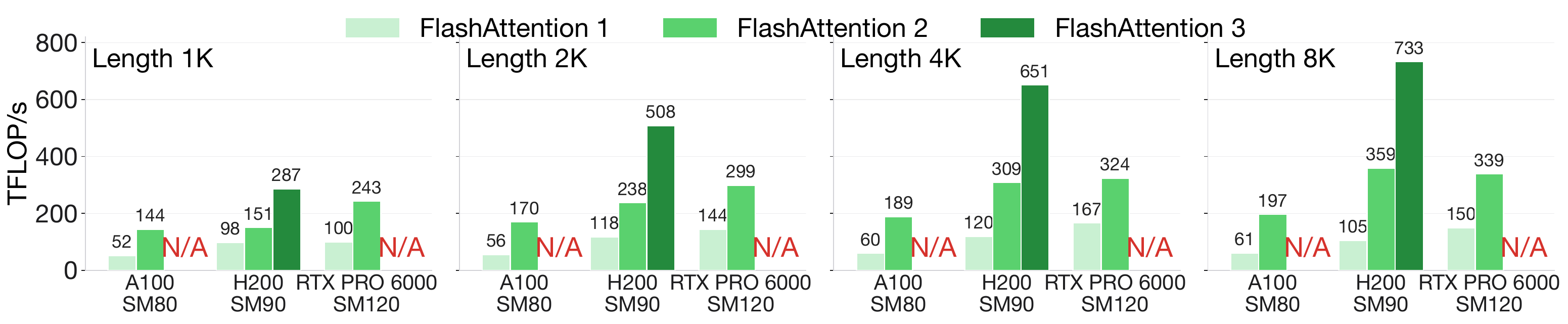}
    \includegraphics[width=\linewidth]{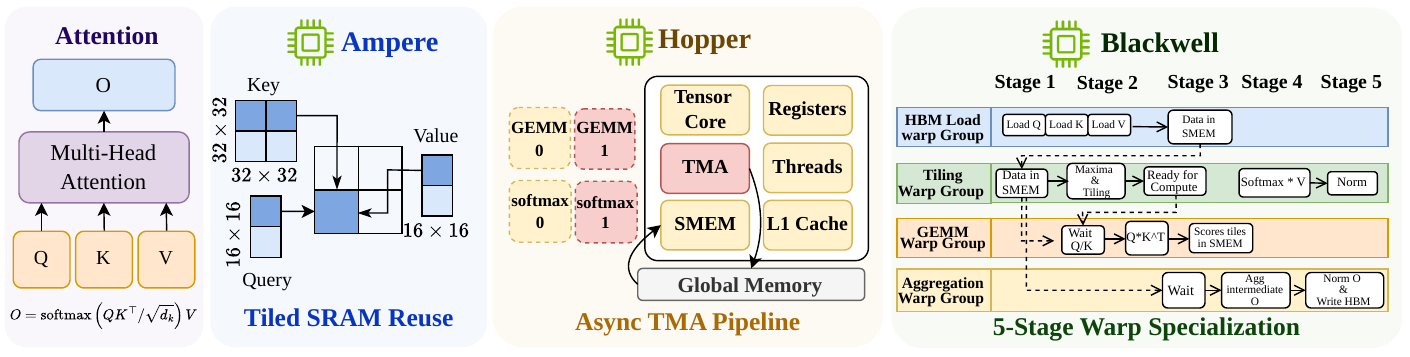}
    \caption{\textbf{Top:} throughput (TFLOP/s) of FA1, FA2, and FA3 on causal forward attention (head dim 128) across A100, H200, and RTX PRO 6000 Blackwell at sequence lengths 1K--8K; \textcolor{red}{N/A} marks unsupported configurations. \textbf{Bottom:} SOTA optimization strategies for self-attention vary across GPU architectures, motivating architecture-aware kernel design.}
    \label{fig:fa-and-intro}
\end{figure}

Existing evaluation benchmarks do not fully answer this question. Most current benchmarks measure whether a model can translate an operator into CUDA or produce a correct kernel under a fixed interface~\cite{ouyang2025kernelbench,li2025tritonbench}. This is useful for measuring code-generation reliability, but it misses a central property of CUDA programming: high-performance kernels are the result of extreme hardware-software co-design. Effective implementations depend on optimization decisions that are tightly coupled to both workload structure and target GPU architecture, and these decisions are not stable across GPU generations. Even within a single workload family, the top panel of \Cref{fig:fa-and-intro} shows that the throughput of FlashAttention variants changes substantially across FlashAttention~V1~\cite{FA1} to FlashAttention~V2~\cite{FA2} and FlashAttention~V3~\cite{FA3} on different GPU architectures. The bottom panel illustrates why: the optimization strategies of FlashAttention move from tiled shared-memory reuse on Ampere to asynchronous pipelines on Hopper and warp-specialized execution on Blackwell, with each variant designed for a specific hardware architecture.

Another limitation is reference quality. Many existing benchmarks compare generated kernels against functional baselines rather than human-expert CUDA implementations. A model can therefore appear successful by producing a correct solution that beats a weak baseline, even if it remains far from human-expert performance and does not recover any architecture-specific optimization strategy. Such evaluation can overestimate progress toward automated CUDA engineering, because it conflates functional synthesis with real deep optimization.

\begin{figure}[t]
    \centering
    \includegraphics[width=\linewidth]{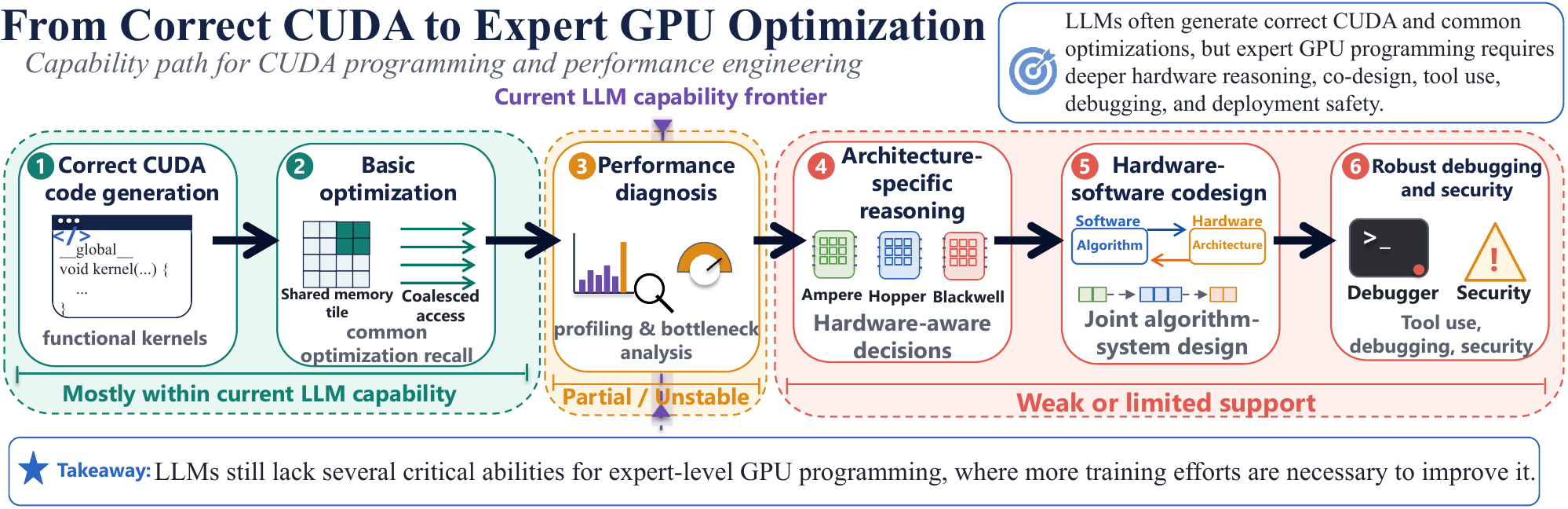}
    \caption{Capability path from correct CUDA generation to expert-level GPU optimization. Current LLMs are strongest at functional kernel generation and common optimization recall, partially reliable at performance diagnosis, but have limited ability in architecture-specific reasoning, hardware-software co-design, and robust debugging or deployment security.}
    \vspace{-5pt}
    \label{fig:cuda-capability-path}
\end{figure}

Together, these limitations point to a capability gap rather than a simple correctness gap. \Cref{fig:cuda-capability-path} decomposes CUDA programming into a path from functional generation and common optimization recall to bottleneck diagnosis, architecture-specific implementation, hardware-software co-design, and robust debugging. \benchmark{} targets these later stages: whether an LLM-based system can produce CUDA implementations that are correct, competitive with human-expert references, specialized to the target GPU, and valid under the semantic constraints of real applications.



We introduce \benchmark, a benchmark for measuring expert-referenced, architecture-aware CUDA optimization by LLM-based systems. Tasks are drawn from curated state-of-the-art CUDA systems beyond ML inference, e.g., cryptographic proving and scientific computing~\cite{cutlassrepo,li2025tritonbench,ICICLE,osama2024certified,fa4,flashinferrepo,thunderkittensrepo,cuszprepo,li2023liberator,TC-GNN,https://doi.org/10.1002/mgea.70028,exachem,MGG}. Each task specifies a target architecture, solution interface, build and execution harness, correctness or semantic validator, expert-level reference, and performance parser through a unified \taskyaml{} metadata schema. Candidate implementations are evaluated against human-expert references rather than naive baselines; architecture-specific subsets target Ampere, Hopper, and Blackwell GPUs; and application tasks require domain-specific semantic validity.

\begin{figure}[t]
    \centering
    \includegraphics[width=\linewidth]{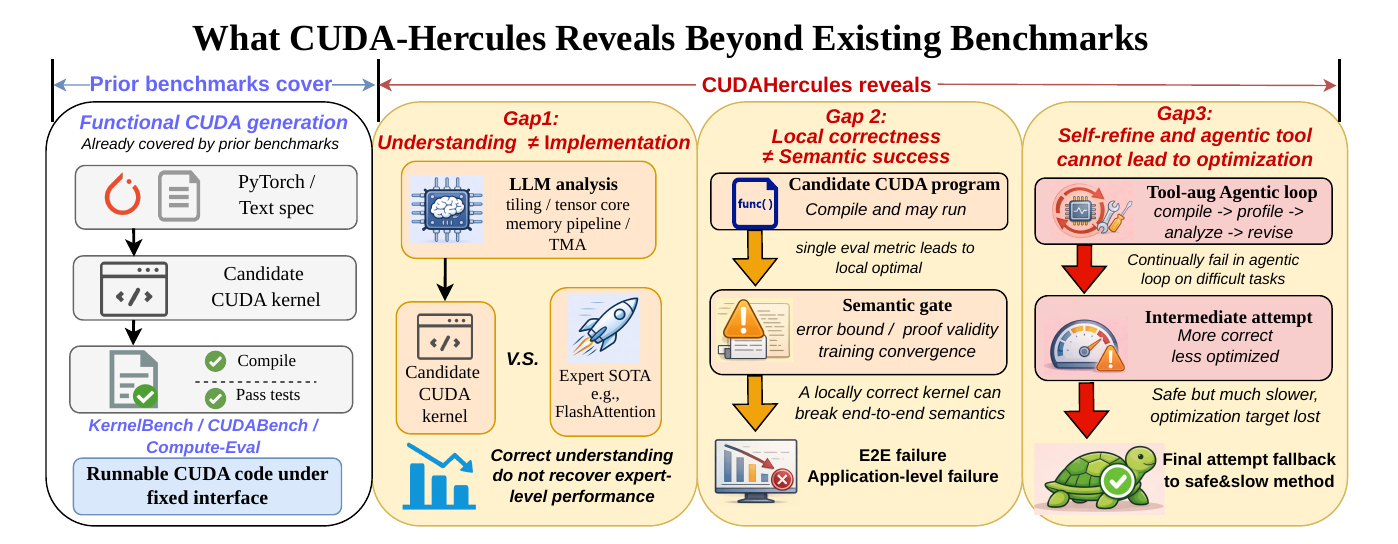}
    \caption{What \benchmark{} reveals beyond existing CUDA-generation benchmarks. Prior benchmarks primarily cover functional CUDA generation under fixed interfaces, while \benchmark{} exposes another three ability gaps for current LLMs.}
    \label{fig:benchmark-gap}
\end{figure}

We evaluate three frontier models, GPT-5.4~\cite{chatgpt}, Claude-Opus-4.6~\cite{claude}, and Qwen3.5-122B-A10B~\cite{qwen3.5}, one domain-specific RL model Kevin32B~\cite{kevin32b}, and two multi-agent methods, CUDAForge~\cite{cudaforge} and StitchCUDA~\cite{stitchcuda}, on \benchmark. As summarized in \Cref{fig:benchmark-gap}, existing CUDA-generation benchmarks mainly cover functional code generation under fixed interfaces, while \benchmark{} exposes three deeper gaps. First, current models can often describe expert-level strategies but fail to realize comparable implementations in code. Second, locally correct kernels can fail under application-level semantic validators such as bounded error or proof verification. Third, agentic refinement can improve correctness while reducing optimization quality, because the search drifts toward slow fallback implementations. These findings suggest that the primary bottleneck is not only CUDA code generation, but also reliably turning optimization knowledge into deployable kernels.

Our main contributions are summarized as follows:
\begin{itemize}[leftmargin=1.5em]
    \item \textbf{An expert-referenced benchmark for LLM-based CUDA programming.} We release \numtasks{} tasks spanning single kernels, modules, full application workloads, and unsolved challenges across multiple domains, with expert-quality references from SOTA systems such as CUTLASS library and FlashAttention-3.
    \item \textbf{Architecture-aware and semantics-gated evaluation.} We evaluate generated CUDA code against expert references under architecture-specific targets across Ampere, Hopper, and Blackwell GPUs, and require application-level semantic validity, including bounded error, convergence, and proof correctness, before assigning performance credit.
    \item \textbf{Architecture-grounded kernel intelligence.} Through evaluation of frontier models and agentic systems, \benchmark{}
    identifies the need for stronger code understanding with hardware architecture-aware implementation, and the risk of degenerative optimization behavior during agentic search. It further argues that effective CUDA acceleration requires joint software-hardware co-design tailored to the practical properties of real GPU operations.
\end{itemize}

\section{Related Work}

\paragraph{GPU kernel generation benchmarks.}
KernelBench~\citep{ouyang2025kernelbench} measures PyTorch-to-CUDA replacement, and recent systems report strong scores on it~\citep{cudaforge,stitchcuda,kevin32b,drkernel,cudaagent}. CUDABench~\citep{zhu2026cudabench} expands text-to-CUDA synthesis with multiple prompt levels and automated correctness checks. TritonBench~\citep{li2025tritonbench} evaluates PyTorch-to-Triton generation, and ComputeEval~\citep{computeeval2025} provides lightweight text-to-CUDA evaluation. These benchmarks establish reproducible functional evaluation, but their reference targets are primarily functional baseline implementations. They do not evaluate generated code against expert-level implementations, and they do not jointly test architecture-specific optimization and application-level semantic validity.

\paragraph{LLM-based automated CUDA programming systems.}
Existing systems use LLMs, agentic search, and tool feedback to generate or optimize CUDA code.  CUDAForge~\citep{cudaforge} incorporates profiling feedback into iterative kernel improvement. StitchCUDA~\citep{stitchcuda} proposes a multi-agent pipeline with rubric-based single-turn reinforcement learning (RL) for end-to-end GPU programming. Both CUDAForge and StitchCUDA note that iterative feedback can produce unstable or degenerative optimization behavior, but their evidence is primarily tied to KernelBench-style functional CUDA tasks, which limits how clearly such behavior can be evaluated against expert-level optimization targets. Kevin32B~\citep{kevin32b} and CUDA Agent~\cite{cudaagent} improve the CUDA programming ability of LLM with multi-turn RL. These systems motivate the evaluation settings in \benchmark: one-shot evaluation measures direct generation, while tool-augmented and iterative-refinement settings measure whether execution feedback and agentic search close the gap to expert implementations. 

\paragraph{Positioning.}
%
\Cref{tab:benchmark-comparison} positions \benchmark{} by evaluation axes. Existing benchmarks primarily test whether a model can produce a runnable CUDA or Triton implementation under a fixed interface. \benchmark{} instead evaluates the stages that remain after basic code generation: matching expert-level references and architecture-specific requirements,
preserving application-level semantics, and reporting compilation, correctness, speed, and agentic-search behavior as separate diagnostics.

\begin{table*}[t]
    \centering
    \footnotesize
    \setlength{\tabcolsep}{2pt}
    \renewcommand{\arraystretch}{1.1}
    \begin{tabularx}{\textwidth}{>{\raggedright\arraybackslash}m{1.85cm} >{\raggedright\arraybackslash}X >{\centering\arraybackslash}m{1.3cm} >{\centering\arraybackslash}m{2.0cm} >{\centering\arraybackslash}m{2.05cm} >{\centering\arraybackslash}m{2.0cm} >{\centering\arraybackslash}m{1.35cm}}
        \toprule
        Benchmark & Setting & Tasks & \shortstack{Multi-domain\\coverage} & \shortstack{Human expert\\reference} & \shortstack{Application-level\\semantics} & \shortstack{Architecture-\\aware} \\
        \midrule
        KernelBench & PyTorch $\rightarrow$ CUDA & 250 & \xmark & \xmark & \xmark & \xmark \\
        CUDABench & Text $\rightarrow$ CUDA & 500 & \cmark & \xmark & \xmark & \xmark \\
        TritonBench-T & PyTorch $\rightarrow$ Triton & 166 & \xmark & \xmark & \xmark & \xmark \\
        ComputeEval & Text $\rightarrow$ CUDA & 566 & \cmark & \xmark & \xmark & \xmark \\
        \rowcolor{gray!12}
        \textbf{\benchmark{}} & Text $\rightarrow$ CUDA & \numtasks{} & \cmark & \cmark & \cmark & \cmark \\
        \bottomrule
    \end{tabularx}
    \caption{Comparison with other GPU kernel generation benchmarks. \benchmark{} is distinguished by expert-level references, architecture-aware evaluation, and application-level semantic validation.}
    \label{tab:benchmark-comparison}
\end{table*}

\section{Benchmark Design}

This section specifies the task organization, execution protocol, and scoring metrics used by \benchmark{}. Each task is a self-contained evaluation unit with a target architecture(SM version), reference solution interface, build and execution harness, correctness or semantic validator, source code of baselines for open-sourced references or binary for closed-sourced references, performance parser, and anti-cheating metadata. Here, anti-cheating metadata refers to validity rules that prevent benchmark-specific shortcuts such as direct reference reuse, disallowed library calls, hard-coded outputs, or timing manipulation.
\begin{figure}[ht]
    \centering
    \includegraphics[width=\linewidth]{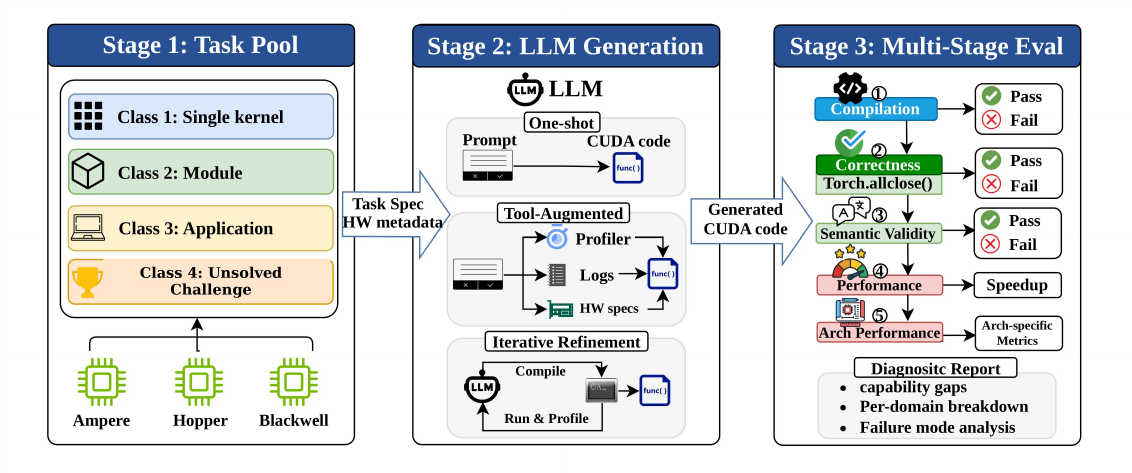}
    \caption{The test workflow of \benchmark.}
    \label{fig:testflow}
\end{figure}

\benchmark{} follows four design requirements:
\begin{enumerate}[leftmargin=1.5em]
    \item \textbf{Use expert references.} Candidate performance is measured against optimized implementations from human-expert level CUDA systems rather than functional naive baselines. FlashAttention-3~\cite{FA3}, for example, provides an expert Hopper reference for self-attention.
    \item \textbf{Encode target hardware.} Task metadata records GPU requirements because optimization strategies depend on architecture-specific instructions, memory pipelines, and scheduling primitives. In the Blackwell\_FP8\_GEMM task, the target is an SM100 GPU, and the expert implementation relies on Blackwell FP8 tensor-core execution, where a generic CUDA GEMM is not equivalent.
    \item \textbf{Cover multiple domains.} The released task set spans ML training and inference, HPC, graph analytics, imaging, GPU compression, formal verification, cryptography, and communication-heavy multi-GPU workloads.
    \item \textbf{Require semantic validation.} Application tasks use domain-specific checks, such as bounded reconstruction error for GPU lossy compression, stable loss decrease for LLM training, and valid proof verification for GPU-based zero-knowledge proving.
\end{enumerate}

\subsection{Task Classes}

We categorize our tasks by the scope of the optimization goal, ranging from local kernel level to application-wise, with specific hardware and software environment setup
, rather than solely the difficulty of the task. The difference stems from our observations that previous benchmarks can become agnostic to detailed HW/SW environments. Because the CUDA implementations behind high-level abstraction (e.g., PyTorch API) vary across HW/SW environments. This can lead to manifest performance discrepancies of reference baselines on the same hardware architecture with different PyTorch versions, or the same PyTorch version on different hardware.

\benchmark{} instead fixes both the target hardware and the expert reference. Class~1 isolates single expert kernels, Class~2 covers module-level or kernel-family optimization, Class~3 evaluates full applications with semantic validators, and Class~4 contains unsolved challenge tasks. Thus, measured speedup reflects the gap to human-expert, architecture-aware CUDA on a specified platform. Detailed task inventories and representative case studies for each class are provided in Appendix~\ref{app:task-catalog}.

Overall, the released task set contains \numtasks{} tasks organized by optimization scope, as summarized in \cref{tab:task-class-summary}. Within Classes~1--2, tasks are additionally bucketed by target architecture. Here, ``general'' denotes tasks that do not require a special architecture-specific instruction subset. Class~3 are evaluated under the NVIDIA RTX PRO 6000 GPU (Blackwell SM120).

\begin{table*}[t]
    \centering
    \footnotesize
    \setlength{\tabcolsep}{6pt}
    \renewcommand{\arraystretch}{1.05}
    \begin{tabular}{llcl}
        \toprule
        Class & Unit of optimization & Tasks & Architecture coverage \\
        \midrule
        Class~1 & Single kernel & \numclassone{} & \cOneGeneral{} general, \cOneHopper{} Hopper, \cOneBlackwell{} Blackwell \\
        Class~2 & Module or kernel family & \numclasstwo{} & \cTwoGeneral{} general, \cTwoHopper{} Hopper, \cTwoBlackwell{} Blackwell \\
        Class~3 & Full application workload & \numclassthree{} & Blackwell evaluation \\
        Class~4 & Unsolved challenge task & \numclassfour{} & Blackwell evaluation \\
        \bottomrule
    \end{tabular}
    \caption{Task classes in \benchmark. Counts report released tasks; architecture buckets apply to Classes~1--2 in the current release.}
    \label{tab:task-class-summary}
    \vspace{-5pt}
\end{table*}

\begin{figure}[ht]
    \centering
    \includegraphics[width=0.9\linewidth]{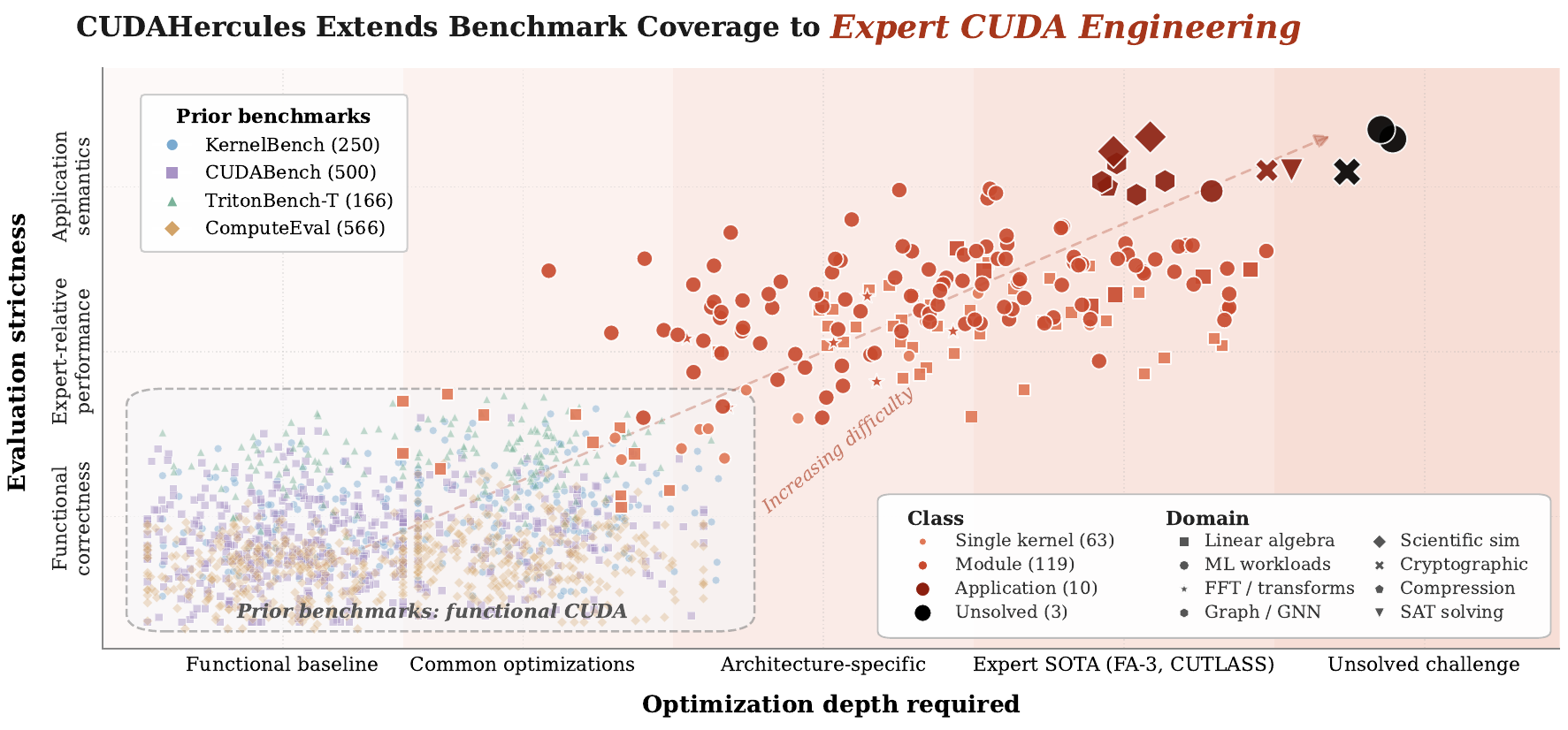}
    \caption{Domain coverage of the \benchmark{} task set.}
    \label{fig:domain-coverage}
\end{figure}

\subsection{Domain Coverage and Task Construction}

A benchmark that only covers ML inference can overfit to one optimization profile. \benchmark{} therefore spans workload groups chosen to exercise qualitatively different optimization regimes. \Cref{fig:domain-coverage} summarizes the domain coverage in the released task set. Detailed descriptions of tasks are provided in Appendix~\ref{app:task-catalog}.

Each task is packaged with a unified metadata file (\taskyaml{}) that specifies hardware requirements, runner backend, build and execute commands, correctness mode, timing parser, source provenance, and anti-cheating rules; Appendix~\ref{app:task-prompt-examples} gives a concrete metadata and prompt example. This separates task specification from execution and lets a unified runner dispatch to backend-specific pipelines. The implementation supports four backends: Makefile-based single-kernel tasks, Python-definition module tasks, full-application tasks, and unsolved challenge tasks. 

\subsection{Evaluation Protocol and Metrics}

We evaluate each model in three settings. \textbf{One-shot} gives the model the task specification, hardware metadata, and few-shot exemplars without execution feedback; an illustrative prompt is shown in Appendix \ref{app:task-prompt-examples}. \textbf{Iterative refinement} uses CUDAForge~\cite{cudaforge} to provide bounded compile, run, and execution feedback over multiple turns. \textbf{Tool-augmented} evaluation exposes build logs, hardware specifications, and GPU profiling tools under the refinement budget, with tool calls selected by the model through LangChain~\cite{LangChain}. We also run strategy-enriched prompt diagnostics, adding high-level expert guidance such as tiling, memory movement, tensor-core use, pipeline organization, or fusion choices while withholding reference code, to test whether failures reflect missing guidance or inability to realize that guidance in CUDA.

As shown in \Cref{fig:testflow}, the runner checks the target GPU requirement, creates an isolated workspace, builds the generated solution, runs the task validator, and measures speed only after validation succeeds. Correctness is backend-defined rather than model-reported: Class~1 uses compiled CUDA harnesses over multiple problem sizes, Class~2 compares generated and reference modules on randomized inputs with task tolerances, and Classes~3--4 use application drivers where validity can depend on semantic checks such as error bounds, convergence criteria, energy tolerance, SAT/UNSAT verdicts, or proof verification. This separation lets the benchmark distinguish compilation, correctness, semantic validity, and performance failures.

For a task set $T$, the compilation rate is the fraction that builds, and the pass rate is the fraction that satisfies the task validator. For a valid task $t$, the expert-relative speed is
\[
s_t=\frac{r_t^{\mathrm{expert}}}{r_t^{\mathrm{gen}}},
\]
where both runtimes are parsed by the task's performance parser; invalid tasks receive zero performance credit. We report mean speed as
\[
|T|^{-1}\sum_{t\in T}\mathbf{1}\{\mathrm{valid}(t)\}s_t
\]
and threshold success as
\[
\fastp{}(p)=|T|^{-1}\sum_{t\in T}\mathbf{1}\{\mathrm{valid}(t)\land s_t\ge p\}.
\]
We use \fastp{}(1.05) as the primary threshold because a 5\% margin reduces sensitivity to system-level timing noise and better reflects a meaningful improvement over the expert reference.

\subsection{Reference Quality and Anti-Cheating Controls}

Because many expert references come from public high-performance CUDA systems, \benchmark{} records source provenance and applies static and procedural anti-cheating controls. In the current release, prompts do not expose the target expert solution, generated code is checked for disallowed library calls and reference reuse, scoring runs in separate workspaces, and passing solutions can be audited for shortcut behavior such as hard-coded outputs or timing manipulation. Audited violations are counted as failures; detailed control rules are provided in Appendix~\ref{app:anti-cheat}.

\section{Experiments}
\label{sec:experiments}

We evaluate GPT-5.4~\cite{chatgpt}, Claude-Opus-4.6~\cite{claude}, Qwen3.5-122B-A10B~\cite{qwen3.5}, and Kevin32B~\cite{kevin32b} on \benchmark. For Classes~1 and~2, we report one-shot pass@1/pass@3, iterative self-refinement with a 10-round revision budget (self-refine@10), and tool-augmented search. The tool-augmented runs are implemented in the CUDAForge framework~\cite{cudaforge}. The ``general'' tasks without special architecture-specific instruction dependencies are measured on RTX PRO 6000 by default, with the Ampere-instruction-dependent subset evaluated on A100; Hopper-targeted tasks are evaluated on H200; and Blackwell-targeted tasks are evaluated on RTX PRO 6000 or B200 depending on the required arch-specific CUDA instruction. Class~3  tasks are evaluated under the StitchCUDA~\cite{stitchcuda}, an end-to-end CUDA programming agentic framework, and all current Class~3 runs are measured on RTX PRO 6000. We do not report pass@N for Class~3 because no evaluated LLM achieves success on any Class~3 task by one-shot generation.
\subsection{Headline Results}

\Cref{tab:general-results} summarizes the results on general subsets of Class~1 (20 tasks) and Class~2 (43 tasks). The strongest one-shot model is GPT-5.4: at pass@3 it reaches 70\% correctness on Class~1 and 81\% on Class~2, with mean expert-relative speed ratios of $0.257\times$ and $0.463\times$, respectively. Claude-Opus-4.6 is consistently second-best in one-shot correctness, while Qwen3.5-122B-A10B falls off sharply on the Class~2 tasks: at pass@3 it compiles 74\% of tasks but solves none of them correctly. Kevin32B solves a small fraction of Class~1 tasks and reaches 20\% correctness on Class~2 under self-refine@10, but its mean speed remains very low.
\begin{table*}[h]
    \centering
    \scriptsize
    \setlength{\tabcolsep}{3pt}
    \renewcommand{\arraystretch}{1.08}
    \resizebox{\textwidth}{!}{%
    \begin{tabular}{llcccccccccccc}
        \toprule
        & & \multicolumn{6}{c}{Class 1 general subset (20)} & \multicolumn{6}{c}{Class 2 general subset (43)} \\
        \cmidrule(lr){3-8} \cmidrule(lr){9-14}
        & & \multicolumn{3}{c}{Outcome} & \multicolumn{3}{c}{\fastp{} threshold} & \multicolumn{3}{c}{Outcome} & \multicolumn{3}{c}{\fastp{} threshold} \\
        \cmidrule(lr){3-5} \cmidrule(lr){6-8} \cmidrule(lr){9-11} \cmidrule(lr){12-14}
        Setting & Model & Compiled & Pass & Mean speed & 0.5 & 0.8 & 1.05 & Compiled & Pass & Mean speed & 0.5 & 0.8 & 1.05 \\
        \midrule
        pass@1 & GPT-5.4 & 70\% & 45\% & $0.231\times$ & 25\% & 10\% & 5\% & 79\% & 72\% & $0.425\times$ & 37\% & 23\% & 14\% \\
        pass@3 & GPT-5.4 & 95\% & 70\% & $0.257\times$ & 30\% & 15\% & 5\% & 86\% & 81\% & $0.463\times$ & 37\% & 28\% & 14\% \\
        pass@1 & Claude Opus 4.6 & 55\% & 45\% & $0.108\times$ & 5\% & 0\% & 0\% & 74\% & 65\% & $0.276\times$ & 19\% & 19\% & 9\% \\
        pass@3 & Claude Opus 4.6 & 70\% & 50\% & $0.124\times$ & 10\% & 0\% & 0\% & 84\% & 58\% & $0.300\times$ & 42\% & 26\% & 9\% \\
        pass@1 & Qwen3.5-122B-A10B & 70\% & 20\% & $0.055\times$ & 5\% & 0\% & 0\% & 40\% & 2\% & $0.005\times$ & 0\% & 0\% & 0\% \\
        pass@3 & Qwen3.5-122B-A10B & 85\% & 35\% & $0.093\times$ & 5\% & 5\% & 0\% & 74\% & 0\% & $0.000\times$ & 0\% & 0\% & 0\% \\
        pass@1 & Kevin32B & 30\% & 10\% & $0.047\times$ & 5\% & 0\% & 0\% & 23\% & 0\% & $0.000\times$ & 0\% & 0\% & 0\% \\
        pass@3 & Kevin32B & 40\% & 15\% & $0.084\times$ & 5\% & 0\% & 0\% & 28\% & 0\% & $0.000\times$ & 0\% & 0\% & 0\% \\
        \midrule
        Self-refine@10 & GPT-5.4 & 85\% & 60\% & $0.309\times$ & 25\% & 20\% & 15\% & 67\% & 56\% & $0.298\times$ & 33\% & 21\% & 7\% \\
        Self-refine@10 & Claude Opus 4.6 & 90\% & 65\% & $0.272\times$ & 20\% & 15\% & 5\% & 70\% & 67\% & $0.451\times$ & 44\% & 30\% & 14\% \\
        Self-refine@10 & Qwen3.5-122B-A10B & 100\% & 90\% & $0.237\times$ & 20\% & 10\% & 5\% & 81\% & 2\% & $0.001\times$ & 0\% & 0\% & 0\% \\
        Self-refine@10 & Kevin32B & 70\% & 20\% & $0.105\times$ & 0\% & 0\% & 0\% & 30\% & 20\% & $0.001\times$ & 0\% & 0\% & 0\% \\
        \midrule
        Tool-augmented & GPT-5.4 & 100\% & 85\% & $0.323\times$ & 25\% & 15\% & 5\% & 100\% & 2\% & $0.020\times$ & 2\% & 2\% & 0\% \\
        Tool-augmented & Claude Opus 4.6 & 100\% & 90\% & $0.403\times$ & 35\% & 20\% & 0\% & 72\% & 47\% & $0.138\times$ & 16\% & 5\% & 2\% \\
        Tool-augmented & Qwen3.5-122B-A10B & 95\% & 85\% & $0.171\times$ & 15\% & 0\% & 0\% & 93\% & 2\% & $0.007\times$ & 0\% & 0\% & 0\% \\
        Tool-augmented & Kevin32B & 30\% & 10\% & $0.125\times$ & 0\% & 0\% & 0\% & 15\% & 0\% & $0.000\times$ & 0\% & 0\% & 0\% \\
        \bottomrule
    \end{tabular}%
    }
    \caption{Results on the general subsets of \benchmark. Compiled reports successfully; Pass reports correctness. Mean speed is the expert runtime divided by the generated runtime. The \fastp{} threshold columns report the fraction of tasks with valid speedup at or above the listed threshold. Self-refine@10 uses a 10-round revision budget with compile and execution feedback; tool-augmented Class~1/2 evaluations use a 20-round budget.}
    \label{tab:general-results}
\end{table*}

Correctness does not translate into expert-level performance. The best one-shot mean speedups are only $0.257\times$ on Class~1 and $0.463\times$ on Class~2, while \fastp{}(1.05) remains low. The \fastp{} thresholds in \Cref{tab:general-results} quantify this distance from human experts: $p=0.5$ is half-speed, $p=0.8$ is near-expert speed, and $p=1.05$ exceeds the expert reference by 5\%. And results show that even \fastp{}(0.5) is still low for all evaluated models, indicating the CUDA optimization ability of LLMs is still far from human experts.

Architecture-specific stress tests are harder still. On Hopper H200, only a few settings reach non-zero \fastp{}(1.05), peaking at 14\% on Class~1 under Claude Opus 4.6 self-refine@10 and 13\% on Class~2 under Claude Opus 4.6 tool augmentation. On Blackwell B200, Claude Opus 4.6 tool augmentation reaches 45\% Class~1 pass but only $0.053\times$ mean speed, no setting achieves non-zero \fastp{}(1.05), and Class~2 remains unsolved. Full results are in Appendix~\ref{app:arch-stress-results}.

\subsection{Understanding-to-Implementation Gap}

The main result in \cref{tab:general-results} is not a ranking among models, but a gap between functional CUDA generation and expert implementation. The strongest models often compile and sometimes pass correctness checks, yet their expert-relative speed remains far below the reference: GPT-5.4 reaches high pass@3 correctness on both Class~1 and Class~2, but its mean speed stays below $0.5\times$, and \fastp{}(1.05) remains low. Qwen3.5-122B-A10B gives an even sharper example on Class~2: it compiles most pass@3 candidates, but none pass correctness. These results show that compilation, API-level behavior, and local correctness are not enough to measure expert-level CUDA ability.

Kevin32B is especially diagnostic. It is a CUDA-specialized model that has already been RL-tuned on KernelBench and reported strong performance there~\cite{kevin32b,ouyang2025kernelbench}. On \benchmark{}, however, it solves only a small fraction of Class~1 tasks, reaches only 20\% correctness on Class~2 under Self-refine@10, and its mean speed remains near zero. This contrast suggests that prior benchmarks can teach models benchmark-specific CUDA generation patterns without training the transferable skills needed to write expert-level, architecture-aware CUDA.

Strategy-enriched prompting also does not close the gap. We tested prompts that inject expert-level optimization guidance, such as intended tiling, memory movement, tensor-core use, and fusion strategy, while still withholding the reference implementation. Models can often restate these strategies or produce plausible design rationales, but the generated code still misses low-level scheduling details, uses the wrong memory hierarchy, fails to coordinate fused stages, or falls back to simpler, correct-but-slow kernels. The bottleneck is therefore not prompt underspecification alone; models must learn to execute high-level optimization plans into correct, architecture-specific, high-performance CUDA code.

\subsection{Agentic Search Helps Inconsistently}

\Cref{tab:general-results} shows that feedback-based workflows can improve Class~1 correctness, but still do not close the performance gap. Claude Opus 4.6 reaches 90\% pass under tool augmentation and Qwen3.5-122B-A10B reaches 90\% under self-refine@10, yet the best Class~1 mean speed ratio is only $0.403\times$ and the best \fastp{}(1.05) is 15\%. Agentic search can therefore repair many functional failures without recovering expert schedules.

\begin{figure*}[t]
    \centering
    \begin{subfigure}{\textwidth}
        \centering
        \includegraphics[width=0.9\textwidth]{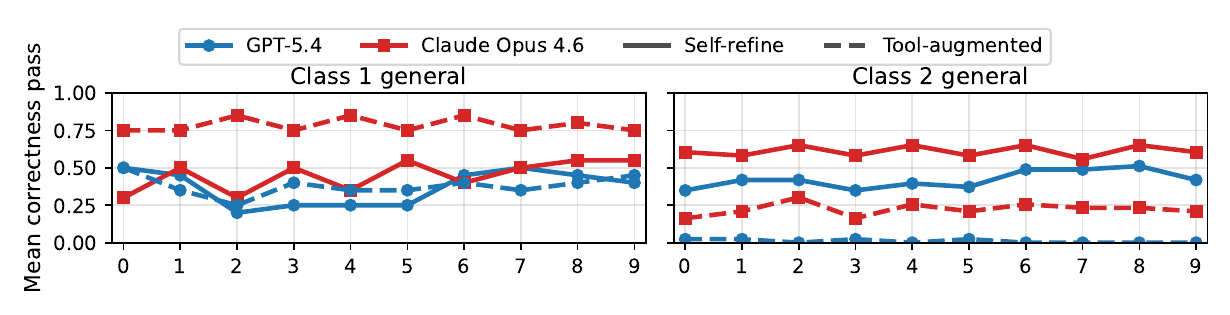}
        \caption{Mean correctness pass.}
        \label{fig:agentic-round-correctness-pass}
    \end{subfigure}
    \vspace{-0.35em}
    \begin{subfigure}{\textwidth}
        \centering
        \includegraphics[width=0.9\textwidth]{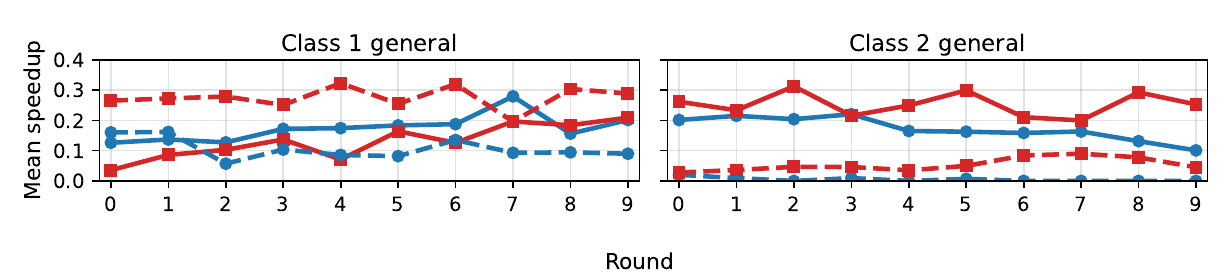}
        \caption{Mean expert-relative speedup.}
        \label{fig:agentic-round-speedup}
    \end{subfigure}
    \caption{Per-round correctness and expert-relative speedup on the Class~1 and Class~2 general subsets under Self-refine@10 and tool augmentation. Incorrect candidates contribute zero.}
    \label{fig:agentic-round-metrics}
\end{figure*}

Class~2 shows the opposite pattern: tool use often degrades outcomes. Claude Opus 4.6 drops from 67\% pass under Self-refine@10 to 47\% under tool augmentation, and GPT-5.4 drops to 2\% under tool augmentation. \Cref{fig:agentic-round-metrics} shows the same behavior round by round: correctness and speed are non-monotonic, and the Class~2 tool-augmented trajectories often collapse. This degenerative behavior was noted by prior agentic systems~\cite{cudaforge,stitchcuda}, but \benchmark{} makes the gap sharper because the search is judged against expert references. This suggests that current models have limited capability to leverage CUDA profiling tools and execution feedback for deep optimization; they often overfit to local diagnostics or preserve compilation while breaking correctness or performance. 

\subsection{Application-Level Semantic Evaluation}

Class~3 is evaluated in the StitchCUDA setting, which is an agentic framework for end-to-end GPU programming with access to tools such as Nsys, NCU, file retrieval, and file editing.  The main empirical result is that application-level semantic gating sharply reduces success. Across the current Class~3 task set, Claude-Opus-4.6 reaches $0.35\times$ on \texttt{cuszp}, $0.74\times$ on \texttt{llmc}, and $0.52\times$ on \texttt{Liberator}. GPT-5.4 only reaches $0.43\times$ on \texttt{Liberator}. Qwen3.5-122B-A10B and Kevin32B do not solve any Class~3 task under the full semantic contract.

The dominant failure mode is not the inability to produce executable CUDA, but the inability to preserve application semantics. GPT-5.4 and Claude-Opus-4.6 both execute full workloads for \texttt{gpumd} and \texttt{exachem\_ccsd\_t}, yet still fail byte-level force checks or energy tolerances. Both models also fail every \texttt{icicle\_zk} round despite repeated successful builds. The resulting picture is stark: once end-to-end semantic validity is enforced, both success rate and performance drop substantially. Class~4 unsolved challenge tasks all remain unsolved in the current evaluation. No evaluated model produces a semantics-valid implementation for any Class~4 task.

\begin{table*}[t]
    \centering
    \footnotesize
    \renewcommand{\arraystretch}{1}
    \begin{tabular}{lcccc}
        \toprule
        & \multicolumn{4}{c}{Model} \\
        \cmidrule(lr){2-5}
        Task & GPT-5.4 & \shortstack{Claude\\Opus 4.6} & \shortstack{Qwen3.5\\122B-A10B} & Kevin32B \\
        \midrule
        \texttt{TC-GNN} & \xmark & \xmark & \xmark & \xmark \\
        \texttt{cuSZp} & \xmark & \cmark~\textbf{$0.35\times$} & \xmark & \xmark \\
        \texttt{llmc} (774M) & \xmark & \cmark~\textbf{$0.74\times$} & \xmark & \xmark \\
        \texttt{Icicle-ZK} & \xmark & \xmark & \xmark & \xmark \\
        \texttt{GPUMD} & \xmark & \xmark & \xmark & \xmark \\
        \texttt{ExaChem} & \xmark & \xmark & \xmark & \xmark \\
        \texttt{ParaFROST} & \xmark & \xmark & \xmark & \xmark \\
        \texttt{Liberator} & \cmark~\textbf{$0.43\times$} & \cmark~\textbf{$0.52\times$} & \xmark & \xmark \\
        \texttt{MGG-AGNN} & \xmark & \xmark & \xmark & \xmark \\
        \texttt{MGG-GCN} & \xmark & \xmark & \xmark & \xmark \\
        \midrule
        Tool calls & 223 & 252 & 196 & 183 \\
        \bottomrule
    \end{tabular}
    \caption{Task-level Class~3 results under the StitchCUDA framework with full tool access. Numeric entries denote semantics-valid expert-relative speedup. `\xmark` means evaluated, but no semantics-valid solution was found. Tool calls count read, write, profile, and planning actions across the Class~3 runs.}
    \label{tab:class3-results}
\end{table*}




\section{Conclusion}

\benchmark{} evaluates LLM-based CUDA generation against expert-level implementations across single kernels, module-level operators, full applications, and unsolved challenge tasks on Ampere, Hopper, and Blackwell GPUs. The results show a persistent ability gap for LLMs between runnable CUDA and expert-level CUDA engineering: frontier models can often compile code and sometimes pass correctness checks, but they rarely recover the hardware-specific optimization strategies needed to match expert references. The gap widens on architecture-specific and application-level tasks. Our results also show that enriched prompts and iterative or tool-augmented feedback do not close this gap. Reliable automated CUDA programming, therefore, requires progress in hardware reasoning, tool-mediated optimization, robust debugging, and application-level semantic validation.

\newpage
\bibliographystyle{plainnat}
\bibliography{references}

\newpage
\appendix

\section{Task Catalog}
\label{app:task-catalog}

This appendix provides a detailed description of the \benchmark~ task set. For each class, we describe the evaluation mechanics, list the full task inventory, and present representative case studies that illustrate the optimization challenges involved.

\subsection{Class 1: Single-Kernel Tasks (\numclassone{} tasks)}
\label{app:class1}

\paragraph{Structure.} Each Class 1 task is a self-contained directory containing:
\begin{itemize}[leftmargin=1.5em]
    \item \texttt{description.txt} --- task specification including function signature, data types, memory layouts, and build instructions. The description is shown to the LLM.
    \item \texttt{solution.h} --- a naive baseline implementation (e.g., a per-thread loop GEMM). The LLM replaces this file.
    \item \texttt{Makefile} --- \texttt{make ref} compiles the expert reference; \texttt{make test} compiles the LLM solution.
    \item \texttt{task.yaml} --- unified metadata: hardware requirements (\texttt{min\_sm}), anti-cheat rules, timing parser regexes, and source provenance.
\end{itemize}

Evaluation compiles the solution via \texttt{make test}, runs it, and parses \texttt{Kernel time:} / \texttt{Ref time:} from stdout to compute speedup. Correctness is verified against the expert reference at multiple problem sizes (small, medium, large). All tasks require the solution to contain \texttt{\_\_global\_\_ void} and kernel launch syntax (\texttt{<<<...>>>}), blocking trivial host-only or library-call submissions.

\paragraph{Architecture distribution.}
Class 1 contains \cOneGeneral{} general tasks, \cOneHopper{} Hopper (SM90) tasks, and \cOneBlackwell{} Blackwell (SM100+) tasks. Here, general means the task does not depend on a special architecture-specific instruction set, although many such tasks still require at least SM80 support and several are drawn from CUTLASS Ampere examples. Hopper tasks include warp-specialized GEMMs, FP8 variants, sparse GEMMs, and DeepGEMM/ThunderKittens kernels. Blackwell tasks cover narrow-precision GEMMs (FP4, NVFP4, MXFP8), MoE GEMMs, and low-latency GQA.

\paragraph{Case study: Fused GEMM + Softmax (\texttt{gemm\_softmax}).}
This task requires implementing a batched GEMM followed by row-wise softmax in a single fused kernel using FP16 data with FP32 accumulation. The function signature takes matrices $A$ (row-major), $B$ (column-major), $C$, $D$, and \texttt{Softmax} output, along with batch strides. For each batch: (1) $D = \alpha \cdot A \cdot B + \beta \cdot C$, then (2) $\text{Softmax}[m,n] = \exp(D[m,n] - \max_n D[m,:]) / \sum_n \exp(D[m,n'] - \max_n D[m,:])$. The naive baseline computes these as separate passes. The expert CUTLASS reference fuses them in a single kernel with shared-memory tiling and online softmax reduction. Matching the reference requires fusing the epilogue with the GEMM mainloop, choosing correct tile shapes, and handling the numerical stability of online softmax within the tiled execution.

\paragraph{Case study: Hopper Warp-Specialized GEMM (\texttt{hopper\_warp\_specialized\_gemm}).}
This task asks for a TF32 GEMM on Hopper using warp specialization. The naive baseline is a per-thread loop with 16$\times$16 blocks. The expert CUTLASS 3.0 reference uses a persistent kernel with producer/consumer warp groups: producer warps issue TMA loads into shared memory via \texttt{cp.async}, while consumer warps execute WGMMA instructions on shared-memory operands. Matching this reference requires understanding Hopper's asynchronous execution model, TMA descriptor setup, and warp-group synchronization --- none of which appear in pre-Hopper CUDA programming.

\subsection{Class 2: Module-Level Tasks (\numclasstwo{} tasks)}
\label{app:class2}

\paragraph{Structure.} Each Class 2 task directory contains:
\begin{itemize}[leftmargin=1.5em]
    \item \texttt{def.py} --- Python definition specifying \texttt{FUNCTION\_SIGNATURE}, \texttt{SCALAR\_ARGS}, \texttt{TOLERANCES}, \texttt{DESCRIPTION}, and functions \texttt{get\_inputs()}, \texttt{get\_outputs()}, \texttt{reference\_fn()}.
    \item \texttt{reference.cu} --- expert CUDA reference calling into the original library (FlashAttention, cuFFT, etc.) via \texttt{extern "C"} linkage.
    \item \texttt{task.yaml} --- metadata including anti-cheat blocked patterns (e.g., \texttt{\#include "flash\_attn"} is blocked).
\end{itemize}

The evaluation pipeline loads \texttt{def.py}, generates random inputs via \texttt{get\_inputs()}, compiles both reference and solution as pybind11 modules via \texttt{torch.utils.cpp\_extension.load()}, checks correctness via \texttt{torch.allclose} (3 trials), and measures performance via CUDA event timing with L2 cache clearing.

\paragraph{Task families.}
Class 2 tasks are organized into five kernel families:
\begin{itemize}[leftmargin=1.5em]
    \item \textbf{Flash Attention (FA2/FA3):} Forward and backward passes across head dimensions (64, 96, 128, 192, 256) and precisions (FP16, BF16), with causal and non-causal variants. General tasks use FA2 (SM80); Hopper tasks use FA3 (SM90) with forward, backward, and split-forward variants.
    \item \textbf{LayerNorm:} Forward and backward passes, plus parallel variants, across 14 hidden dimensions (256--8192). Reference from flash-attention's layer\_norm kernels.
    \item \textbf{FFT:} Complex-to-complex (C2C) and real-to-complex (R2C) transforms in 1D/2D/3D, across sizes ($2^{10}$--$2^{18}$) and precisions (FP32, FP64). Reference from VkFFT.
    \item \textbf{SageAttention:} Quantized attention with INT8 QK and FP16 PV (Hopper, SM90), and FP4 BlockScaled attention with NVFP4 MMA (Blackwell, SM120).
    \item \textbf{ThunderKittens:} Tile-primitive kernels including MHA, linear attention, Mamba2, and Hedgehog variants (Hopper/Blackwell).
\end{itemize}

Class 2 contains \cTwoGeneral{} general tasks, \cTwoHopper{} Hopper tasks, and \cTwoBlackwell{} Blackwell tasks.

\paragraph{Case study: Flash Attention Backward (\texttt{flash\_attn\_bwd\_hdim128\_bf16}).}
This task requires implementing the FA2 backward pass for BF16 with head dimension 128. Given upstream gradient $dO$ and saved forward tensors ($Q, K, V, O, \text{lse}$), the kernel must compute $dQ, dK, dV$ without materializing the full $S \times S$ attention matrix. The tiled algorithm iterates over K/V tiles nested with Q tiles, recomputing attention probabilities from saved LSE values. The expert FlashAttention reference achieves this with a multi-pass tiled kernel that fuses $dQ$, $dK$, $dV$ accumulation. Tolerance is 5\% (atol and rtol) due to BF16 numerical error in the backward pass. Matching the reference requires understanding online softmax gradients, tiled recomputation, and shared-memory staging for the double loop over Q and K/V tiles.

\paragraph{Case study: SageAttention INT8 (\texttt{sageattn\_qk\_int8\_pv\_fp16\_hdim128}).}
This Hopper task implements quantized attention where $Q$ and $K$ are pre-quantized to INT8 with per-warp scale factors, while $V$ remains in FP16. The kernel computes: (1) $S_{\text{int32}} = Q_{\text{int8}} \cdot K_{\text{int8}}^T$ using INT8 tensor core MMA, (2) dequantize $S_{\text{float}} = S_{\text{int32}} \cdot q_{\text{scale}} \cdot k_{\text{scale}}$, (3) softmax, (4) $O = P \cdot V$ using FP16 MMA. The challenge is coordinating two different MMA datapaths (INT8 for QK, FP16 for PV) within a single fused kernel while managing per-warp quantization scales across shared memory.

\paragraph{Case study: 1D FFT (\texttt{fft\_c2c\_1d\_1024\_fp32}).}
This task requires a 1D complex-to-complex FFT of size 1024 batched over 1024 independent transforms. The reference uses cuFFT, a runtime code-generated FFT library that emits architecture-specific kernels. Anti-cheat rules block both \texttt{cufftExec} and \texttt{VkFFTAppend} calls. The LLM must implement the Cooley--Tukey butterfly from scratch, choosing radix decomposition, shared-memory bank-conflict avoidance, and thread-to-element mapping without access to any FFT library.

\subsection{Class 3: Application-Level Tasks (\numclassthree{} tasks)}
\label{app:class3}

Class 3 tasks are drawn from complete application codebases. The LLM optimizes all kernel files in the application; correctness is verified via application-level checks (loss convergence, output matching, error bounds) rather than tensor comparison alone.  A model must therefore produce architecture-appropriate optimizations for the same application codebase under specific hardware profiles. All current Class~3 evaluations are run on RTX PRO 6000.

\paragraph{Task overview.}
\Cref{tab:class3-tasks} summarizes all \numclassthree{} Class 3 tasks.

\begin{table*}[t]
\centering
\footnotesize
\setlength{\tabcolsep}{3pt}
\renewcommand{\arraystretch}{1.1}
\begin{tabularx}{\textwidth}{>{\raggedright\arraybackslash}p{2.0cm} >{\raggedright\arraybackslash}p{2.5cm} >{\raggedright\arraybackslash}X >{\raggedright\arraybackslash}p{3.2cm} >{\centering\arraybackslash}p{0.8cm}}
\toprule
Task & Domain & Description & Semantic constraint & Files \\
\midrule
\texttt{cuszp} & Lossy compression & Error-bounded GPU compression/decompression across 1D/2D/3D and FP32/FP64 & Every decompressed value within prescribed error bound & 6 \\
\texttt{exachem} & Quantum chemistry & CCSD(T) perturbative triples with FP64 tensor cores & Energy within $10^{-6}$ relative tolerance & 3 \\
\texttt{gpumd} & Molecular dynamics & End-to-end MD across Lennard-Jones, Tersoff, and Coulomb/Ewald force models & Energy drift $< 0.05$ over simulation & 1 \\
\texttt{icicle\_zk} & ZK cryptography & NTT and MSM on BN254 for zero-knowledge proofs & Exact match against CPU reference & 1 dir \\
\texttt{liberator} & Graph processing & BFS, CC, SSSP, PageRank on Friendster graph under GPU memory limits & Correct output at all memory limits & 1 \\
\texttt{llmc} & LLM training & GPT-2 774M training pipeline (attention, layernorm, GELU, AdamW) & Loss convergence: final loss $< 8.0$ & 8 \\
\texttt{mgg\_agnn} & GNN training & Attention-based GNN with cosine-similarity edge weights & Training accuracy convergence & 1 \\
\texttt{mgg\_gcn} & Multi-GPU GNN & 4-GPU GCN with NVSHMEM communication-computation overlap & Correct aggregation across GPUs & 1 \\
\texttt{parafrost\_sat} & SAT solving & GPU-accelerated CDCL SAT solver with inprocessing & Correct SAT/UNSAT verdict & 13 \\
\texttt{tcgnn\_gcn} & GNN training & Tensor Core SpMM for GCN neighbor aggregation & Training accuracy convergence & 1 \\
\bottomrule
\end{tabularx}
\caption{Class 3 application-level tasks. ``Files'' indicates the number of solution files the LLM must optimize.}
\label{tab:class3-tasks}
\end{table*}

\paragraph{Case study: GPT-2 Training (\texttt{llmc}).}
This task optimizes end-to-end GPT-2 774M training. The LLM receives 8 kernel modules: attention (permute, causal softmax, unpermute), layernorm (forward, fused residual+layernorm), GELU, token/position embedding, fused cross-entropy classifier, AdamW optimizer, gradient norm, and cuBLAS matmul wrappers. Training runs 20 steps on TinyShakespeare with BF16 precision. Correctness requires that loss decreases monotonically and the final loss is below 8.0. Performance is total training time. Unlike Class 2 tasks where each kernel is evaluated independently, here the LLM must co-optimize 8 interacting kernel modules under a global training correctness constraint.

\paragraph{Case study: Error-Bounded Compression (\texttt{cuszp}).}
This task optimizes the cuSZp GPU lossy compressor across 6 variants (1D/2D/3D $\times$ FP32/FP64) at three error-bound levels (relative error $10^{-2}$, $10^{-3}$, $10^{-4}$). Correctness is not just ``output matches reference'' but a hard guarantee: every decompressed value must lie within the prescribed error bound of the original. The test suite runs 54 configurations (6 variants $\times$ 3 modes $\times$ 3 error bounds) on 2\,GB datasets. A kernel that is fast but violates even one error bound fails the entire configuration.

\paragraph{Case study: GPU SAT Solver (\texttt{parafrost\_sat}).}
This task provides 13 kernel and support files from the ParaFROST SAT solver, implementing GPU-accelerated inprocessing for conflict-driven clause learning. The GPU kernels handle occurrence table construction, variable elimination (3-phase: candidate computation, prefix scan, resolvent generation), clause subsumption, blocked clause elimination, and memory recycling. Correctness is binary: the solver must produce the correct SAT/UNSAT verdict on each test instance. This is one of the most structurally complex tasks in the benchmark, requiring the LLM to reason about multi-kernel dataflow, dynamic memory management, and formal correctness of clause-learning invariants.

\subsection{Class 4: Unsolved Challenge Tasks (\numclassfour{} tasks)}
\label{app:class4}

Class 4 tasks target problems where strong implementations exist in higher-level abstractions but no equivalent hand-written CUDA has been demonstrated. These are reported as a separate unsolved challenge track.

\paragraph{FlashAttention-4 Forward (\texttt{flash\_attn4\_fwd}).}
This task asks for a hand-written CUDA implementation of the FA4 forward pass on Blackwell (SM100), without using the CuTe DSL. The FA4 algorithm introduces four key innovations over FA3: (1) conditional softmax rescaling that skips ${\sim}90\%$ of rescaling operations by only triggering when the row-max exceeds the previous maximum by $\tau = \log_2(256) = 8.0$; (2) software exponential emulation using a degree-3 polynomial on FMA units, reserving the hardware EX2 unit for the ${\sim}10$--$25\%$ of entries needing higher precision; (3) 5-way warp specialization with concurrent load, MMA, softmax, correction, and epilogue warps; and (4) Blackwell-specific hardware including TMEM (256\,KB programmer-managed L1), \texttt{tcgen05.mma} (fully asynchronous 128$\times$128 MMA), and 2-CTA MMA mode.

The task provides the full FA4 paper, the official CuTe DSL reference implementation (2842 lines), and helper modules for TMA, WGMMA, and TMEM. The LLM must translate these abstractions into raw CUDA/PTX. Evaluation tests multiple configurations (head dimensions 64/128/256, sequence lengths 1K--32K, causal/non-causal, GQA) and reports TFLOPS against the CuTe baseline. This task requires SM100 hardware.

\paragraph{FlashAttention-4 Backward (\texttt{flash\_attn4\_bwd}).}
This task asks for a hand-written CUDA/PTX implementation of the FA4 backward pass on Blackwell (SM100), without using CuTe DSL or existing FlashAttention backward kernels. Given $Q$, $K$, $V$, saved forward output $O$, upstream gradient $dO$, and saved log-sum-exp values, the implementation must compute $dQ$, $dK$, and $dV$ for BF16 attention with FP32 accumulation.

The task requires recomputing tiled attention scores without materializing the full attention matrix, reconstructing softmax probabilities from saved LSE, computing $dV = P^T dO$ and $dS = P \cdot (dP - \mathrm{rowsum}(dO \cdot O))$, and accumulating $dQ$ and $dK$ through tiled QK recomputation. It must support causal and non-causal masking, GQA, head dimensions 64/128/256, and sequence lengths up to 32K in the full challenge setting. Correctness is checked against PyTorch SDPA autograd with numerical tolerances, and performance is measured as total backward time. This task requires SM100 features including TMA, \texttt{tcgen05.mma}, TMEM accumulators, and warp-specialized producer/consumer pipelines.

\paragraph{Groth16 ZK Prover (\texttt{groth16\_prover}).}
This task asks for a complete GPU Groth16 zero-knowledge proof prover in hand-written CUDA. Groth16 is the most widely deployed zkSNARK (used in Zcash, Filecoin, many L2 rollups). The proving algorithm requires: (1) 6--7 NTTs (forward and inverse) over the BN254 scalar field on domains of size $2^{20}$--$2^{26}$; (2) 5 multi-scalar multiplications on BN254 G1 and G2 curves; and (3) full 254-bit modular and elliptic curve arithmetic (Montgomery multiplication, projective point addition/doubling, quadratic extension field operations for G2).

No hand-written CUDA Groth16 prover exists; all current implementations (Icicle, cuZK, bellman) rely on library primitives. The LLM must implement NTT, MSM (Pippenger bucket method), and BN254 arithmetic from scratch, then orchestrate them into the full proving pipeline. Correctness is verified by checking that the generated proof passes the Groth16 pairing-based verification equation. Performance is compared against the Icicle GPU prover. This task requires SM80+ hardware.

\section{Task Metadata and Prompt Examples}
\label{app:task-prompt-examples}

This section shows the concrete artifacts used to instantiate a task. The example is adapted from the Class~1 \texttt{gemm\_softmax} task and omits only repository-local path details that are not shown to the model. The task name is shown without the internal numeric prefix used in early development.

\paragraph{Example \taskyaml{} metadata.}
The metadata file records the target GPU requirement, runner backend, build and execution commands, correctness policy, performance parser, provenance, and anti-cheating checks. The benchmark runner reads this file before constructing a model prompt or executing a generated solution.

\begingroup
\scriptsize
\begin{verbatim}
schema_version: 1
task_id: class1/general/gemm_softmax
name: gemm_softmax
task_class: 1
domain: ml
tags:
  - gemm
  - softmax
  - fusion
hardware:
  min_sm: 80
  tested_sms: [80]
  required_features: []
runner:
  backend: class1_make
  workdir: tasks/class1/general/gemm_softmax
  timeout_sec: 180
  env:
    KH_BENCHMARK: "20"
  solution_file: solution.h
build:
  cmd: make test
  clean_cmd: make clean
execute:
  cmd: ./test
  success:
    exit_code: 0
    stdout_regex: Passed
correctness:
  mode: stdout_or_exit
  tolerances:
    atol: 0.01
    rtol: 0.01
performance:
  enabled: true
  parser:
    kernel_time_ms_regex: "Kernel time:\\s*([0-9.]+)\\s*ms"
    ref_time_ms_regex: "Ref time:\\s*([0-9.]+)\\s*ms"
  metric: speedup_ref_over_sol
source:
  repo: NVIDIA/cutlass
  files:
    - examples/35_gemm_softmax
  commit: "<recorded commit hash>"
  license: BSD-3-Clause
anti_cheat:
  blocked_patterns: []
  required_patterns:
    - "__global__\\s+void"
    - "<<<.*>>>"
\end{verbatim}
\endgroup

\paragraph{Example one-shot prompt.}
The one-shot prompt combines the task description, hardware metadata, solution interface, and build and scoring rules. When few-shot exemplars are enabled, they are appended after the task-specific block shown below. The prompt does not expose the expert implementation that is used for scoring.

\begingroup
\scriptsize
\begin{verbatim}
System:
You are an expert CUDA programmer. Write a correct and high-performance CUDA
implementation for the requested task. Return only the replacement source file.

User:
Task: gemm_softmax
Class: single-kernel CUDA optimization
Target GPU: NVIDIA SM80 or newer
Source file to replace: solution.h

Goal:
Implement a CUDA solution for a fused batched GEMM followed by row-wise softmax.
The evaluator will compile your solution with `make test`, run `./test`, check
correctness against an expert CUDA reference, and parse runtime from stdout.

Interface:
  You must provide the function and kernel definitions expected by solution.h.
  The solution must include at least one __global__ kernel and launch it with
  CUDA <<<...>>> syntax. Do not change the public function signature.

Inputs and semantics:
  For each batch, compute D = alpha * A * B + beta * C, then compute
  Softmax[m, n] = exp(D[m, n] - max(D[m, :])) /
                  sum_j exp(D[m, j] - max(D[m, :])).
  A is row-major, B is column-major, and the output softmax tensor is row-major.

Hardware and optimization requirements:
  The target GPU supports SM80 features. A strong solution should exploit
  tiled matrix multiplication, shared-memory reuse, numerically stable softmax
  reduction, and fusion between the GEMM epilogue and softmax computation.

Correctness and scoring:
  The program must print "Passed" under the benchmark harness. Correctness is
  checked at multiple problem sizes with atol=1e-2 and rtol=1e-2. Performance
  is scored as expert_runtime / generated_runtime, using the harness-reported
  "Ref time:" and "Kernel time:" values. Incorrect or non-compiling solutions
  receive zero performance credit.

Restrictions:
  Do not call cuBLAS, CUTLASS, Thrust, or any external GEMM/softmax library.
  Do not hard-code outputs, inspect timing loops, bypass the test harness, or
  depend on fixed pointer identities. The solution must be general for the
  problem sizes used by the evaluator.

Return:
  Only the complete contents of solution.h.
\end{verbatim}
\endgroup
\newpage
\section{Detailed Architecture Stress-Test Results}
\label{app:arch-stress-results}

\Cref{tab:arch-stress-results} gives the full breakdown for the architecture-specific Hopper and Blackwell stress tests summarized in the main text. We report pass count, pass rate, mean expert-relative speed, and \fastp{}(1.05) after applying the benchmark's validity checks. The Hopper rows combine the architecture-specific Class~1 and Class~2 slices evaluated on H200.
 
On Hopper Class~1, refinement is the most effective strategy: GPT-5.4 reaches 57\% pass rate under both refine@10 and tool augmentation, while Claude Opus 4.6 peaks at 48\% under tool augmentation. Claude Opus 4.6 reaches the highest Class~1 \fastp{}(1.05), with 14\% under refine@10 and 5\% under pass@3. On Hopper Class~2, GPT-5.4 is stronger under single-sample and few-sample settings (33\% pass@1, 34\% pass@3), whereas Claude Opus 4.6 benefits substantially from tool augmentation, reaching 47\% pass rate with a mean speed of 
0.323× and \fastp{}(1.05) of 13\%. Across both Hopper slices, Qwen3.5-122B-A10B and Kevin32B solve no tasks.

The Blackwell rows separate the 22-task Class~1 SM100a subset from the 12-task Class~2 subset. On Blackwell Class~1, Claude Opus 4.6 obtains the highest pass rate, reaching 45\% under tool augmentation, while GPT-5.4 reaches 41\% under refinement. However, both remain far below the expert references: all \fastp{}(1.05) values are 0, and the best mean speed is only $0.053\times$. The Blackwell Class~2 subset is fully unsolved across all models.

\begin{table}[htbp]
    \centering
    \scriptsize
    \setlength{\tabcolsep}{4pt}
    \renewcommand{\arraystretch}{1.1}
    \resizebox{\textwidth}{!}{%
    \begin{tabular}{lllccccc}
        \toprule
        Target & Model & Setting & Tasks & Pass & Pass rate & Mean speed & \fastp{}(1.05)  \\
        \midrule
        \multirow{10}{*}{\shortstack[l]{Hopper H200\\C1}} 
        & \multirow{4}{*}{GPT-5.4} & pass@1 & 21 & 3 & 14\% & $0.012\times$ & 0\% \\
        & & pass@3 & 21 & 2 & 10\% & $0.004\times$ & 0\%  \\
        & & refine@10 & 21 & 12 & 57\% & $0.053\times$ & 0\%  \\
        & & tool-augmented & 21 & 12 & 57\% & $0.081\times$ & 0\%  \\
        \cmidrule(l){2-8}
        & \multirow{4}{*}{Claude Opus 4.6} & pass@1 & 21 & 5 & 24\% & $0.045\times$ & 0\%  \\
        & & pass@3 & 21 & 5 & 24\% & $0.076\times$ & 5\%  \\
        & & refine@10 & 21 & 8 & 38\% & $0.090\times$ & 14\%  \\
        & & tool-augmented & 21 & 10 & 48\% & $0.031\times$ & 0\%  \\
        \cmidrule(l){2-8}
        & Qwen3.5-122B-A10B & all evaluated & 21 & 0 & 0\% & $0.000\times$ & 0\%  \\
        \cmidrule(l){2-8}
        & Kevin32B & all evaluated & 21 & 0 & 0\% & $0.000\times$ & 0\%  \\
        \midrule
        \multirow{10}{*}{\shortstack[l]{Hopper H200\\C2}} 
        & \multirow{4}{*}{GPT-5.4} & pass@1 & 64 & 21 & 33\% & $0.196\times$ & 6\%  \\
        & & pass@3 & 64 & 22 & 34\% & $0.166\times$ & 6\%  \\
        & & refine@10 & 64 & 15 & 23\% & $0.122\times$ & 5\%  \\
        & & tool-augmented & 64 & 16 & 25\% & $0.109\times$ & 5\%  \\
        \cmidrule(l){2-8}
        & \multirow{4}{*}{Claude Opus 4.6} & pass@1 & 64 & 11 & 17\% & $0.053\times$ & 0\%  \\
        & & pass@3 & 64 & 20 & 31\% & $0.211\times$ & 11\%  \\
        & & refine@10 & 64 & 7 & 11\% & $0.098\times$ & 2\%  \\
        & & tool-augmented & 64 & 30 & 47\% & $0.323\times$ & 13\%  \\
        \cmidrule(l){2-8}
        & Qwen3.5-122B-A10B & all evaluated & 64 & 0 & 0\% & $0.000\times$ & 0\%  \\
        \cmidrule(l){2-8}
        & Kevin32B & all evaluated & 64 & 0 & 0\% & $0.000\times$ & 0\%  \\
        \midrule
        \multirow{10}{*}{\shortstack[l]{Blackwell B200\\C1}} & \multirow{4}{*}{GPT-5.4} & pass@1 & 22 & 0 & 0\% & $0.000\times$ & 0\%  \\
        & & pass@3 & 22 & 1 & 5\% & $0.005\times$ & 0\%  \\
        & & refine@10 & 22 & 9 & 41\% & $0.052\times$ & 0\%  \\
        & & tool-augmented & 22 & 1 & 5\% & $0.009\times$ & 0\%  \\
        \cmidrule(l){2-8}
        & \multirow{4}{*}{Claude Opus 4.6} & pass@1 & 22 & 9 & 41\% & $0.015\times$ & 0\%  \\
        & & pass@3 & 22 & 9 & 41\% & $0.018\times$ & 0\%  \\
        & & refine@10 & 22 & 9 & 41\% & $0.043\times$ & 0\%  \\
        & & tool-augmented & 22 & 10 & 45\% & $0.053\times$ & 0\%  \\
        \cmidrule(l){2-8}
        & Qwen3.5-122B-A10B & all evaluated & 22 & 0 & 0\% & $0.000\times$ & 0\%  \\
        \cmidrule(l){2-8}
        & Kevin32B & all evaluated & 22 & 0 & 0\% & $0.000\times$ & 0\%  \\
        \midrule
        \multirow{4}{*}{\shortstack[l]{Blackwell B200\\C2}} & GPT-5.4 & all evaluated & 12 & 0 & 0\% & $0.000\times$ & 0\%  \\
        & Claude Opus 4.6 & all evaluated & 12 & 0 & 0\% & $0.000\times$ & 0\%  \\
        & Qwen3.5-122B-A10B & all evaluated & 12 & 0 & 0\% & $0.000\times$ & 0\%  \\
        & Kevin32B & all evaluated & 12 & 0 & 0\% & $0.000\times$ & 0\%  \\
        \bottomrule
    \end{tabular}%
    }
    \caption{Hopper and Blackwell architecture stress-test results. Hopper C1 and C2 report the Class~1 and Class~2 architecture-specific slices separately. Blackwell C1 reports the 22-task SM100a. Blackwell C2 reports the 12-task SM100a subset.}
    \label{tab:arch-stress-results}
\end{table}

\section{Reference Quality and Anti-Cheating Controls}
\label{app:anti-cheat}

\benchmark{} uses the following controls to reduce contamination, library bypass, direct reference reuse, and benchmark-specific shortcut behavior.

\begin{itemize}[leftmargin=1.5em]
    \item \textbf{Source provenance and prompt construction.} Each task records source repository, file, commit, and license metadata when available. Prompts are built from task descriptions, interfaces, metadata, solution templates, and selected context files, rather than by directly handing the model the expert kernel as the target solution.
    \item \textbf{Static anti-cheat checks.} Task metadata and global rules block disallowed library calls, reference includes, shortcut APIs, or required-kernel violations before a solution is scored.
    \item \textbf{Workspace and reference separation.} Evaluators run submitted code in temporary workspaces and separate scoring references or private application baselines from the generated solution path, so a candidate is measured through the task harness rather than by directly reusing the scorer.
    \item \textbf{Post-evaluation hacking audit.} Passing solutions can be audited for benchmark-specific shortcut behavior such as pointer-identity caches, reference re-imports, hard-coded outputs, timing-loop manipulation, or suspicious speedups; audited hacking cases are counted as failures in the reported results.
\end{itemize}

\end{document}